\documentclass[sigconf]{acmart}

\usepackage{bbding}
\usepackage{multirow}
\usepackage{booktabs}
\usepackage{algorithm}
\usepackage{algorithmic}
\usepackage{booktabs}
\usepackage{tabularx}
\newcolumntype{Y}{>{\centering\arraybackslash}X}
\usepackage{array}
\usepackage{graphicx}

\usepackage{makecell}
\usepackage{subcaption}

\AtBeginDocument{%
  }

\copyrightyear{2026}
\acmYear{2026}
\setcopyright{cc}
\setcctype{by}
\acmConference[MM '26]{Proceedings of the 34th ACM International Conference on Multimedia}{November 10--14, 2026}{Rio de Janeiro, Brazil}
\acmBooktitle{Proceedings of the 34th ACM International Conference on Multimedia (MM '26), November 10--14, 2026, Rio de Janeiro, Brazil}
\acmDOI{10.1145/3767308.3834941}
\acmISBN{979-8-4007-2213-4/2026/11}
\begin{document}

\title{Emo-DVS: A Multimodal Benchmark for
Privacy-Aware Emotion Recognition with Event Cameras}

\author{Jiaqi Chen}
\affiliation{%
  \institution{Beijing Institute of Technology}
  \city{Beijing}
  \country{China}
}
\email{jq\_chen924@163.com}

\author{Qinfu Xu}
\affiliation{%
  \institution{Beijing Institute of Technology}
  \city{Beijing}
  \country{China}
}
\email{xqfupc@163.com}

\author{Hao Zhuang}
\affiliation{%
  \institution{Beijing Institute of Technology}
  \city{Beijing}
  \country{China}
}
\email{haozhuang@bit.edu.cn}

\author{Liyuan Pan}
\authornote{Corresponding author.}
\affiliation{%
  \institution{Beijing Institute of Technology}
  \city{Beijing}
  \country{China}
}
\email{liyuan.pan@bit.edu.cn}

\begin{abstract}

Emotion analysis is a fundamental task in computer vision, but its practical deployment remains constrained by the privacy risks inherent to conventional RGB cameras. Bio-inspired event cameras present a promising hardware-level solution because they capture asynchronous brightness changes, thereby reducing exposure of facial identity details while leveraging high dynamic range for robust perception under challenging illumination conditions. Despite these advantages, existing event-based methods struggle in complex real-world settings due to limited dataset scales, simple acquisition conditions, and reliance on single-modality visual cues. To address these, we establish a challenging tri-modal benchmark with event, audio, and text modalities and propose the Information-Guided Gated Fusion (IGF) framework, which first pre-trains an event encoder on the FAU subset of Emo-DVS to capture fine-grained facial dynamics, then employs adaptive modality gating to suppress modality-specific noise, and finally leverages mutual information maximization to align robust cross-modal representations. To alleviate data scarcity, we introduce Emo-DVS, the first large-scale event-based emotion analysis dataset, which couples dynamic illumination with the Facial Action Unit (FAU) subset and emotion subset. Extensive experiments demonstrate that IGF achieves state-of-the-art performance. \href{https://github.com/Typistchen/Emo-DVS}{[Code]}

\end{abstract}




\maketitle

\section{Introduction}

\begin{figure}
    \centering
  \includegraphics[scale=1.17]{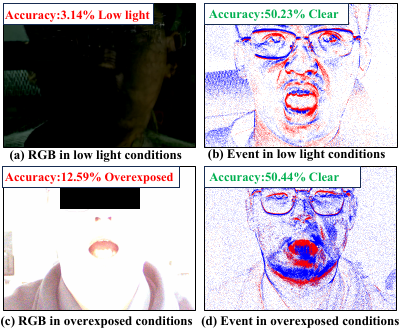}
    \caption{ Limitations of RGB sensors and the advantages of event cameras under challenging illumination scenarios. (a) (c) Traditional RGB sensors struggle to extract textural and features under low-light and overexposure. (b) (d) In contrast, event cameras capture clear facial details under these lighting conditions.}
    \label{fig:intrduction}
\vspace{-5pt}
\end{figure}

Emotion analysis serves as a core task in computer vision with broad applications in psychological assessment \cite{zhao2025context, lee2025v}, human-computer interaction \cite{kim2025contextface, yu2025cross}, and intelligent security surveillance \cite{wang2023rethinking, aquib2025decoupling}. However, most existing methods rely on RGB sensors that capture identifiable facial details and static textures, inevitably leading to privacy leakage concerns in real-world deployments \cite{li2025fed, kim2025privacy}. 
With growing concerns regarding privacy protection, extracting facial features via RGB cameras restricts the practical deployment of emotion analysis systems \cite{ma2025multimodal, cabacas2025enhancing}.

Unlike RGB sensors, bio-inspired event cameras \cite{zhang2023blink, wang2025cs3d} only record asynchronous brightness changes, emerging as an  alternative to address privacy leakage at the hardware level. This  paradigm naturally filters out static background textures and detailed facial identities \cite{becattini2025neuromorphic}, achieving privacy preservation. Furthermore, as shown in Fig. \ref{fig:intrduction}, due to their high dynamic range (HDR), event cameras exhibit robust perception capabilities under various illumination conditions, not only in low-light environments \cite{chen2023multi, adra2025event}, but also under illumination overexposure \cite{savchenko2025leveraging}, where RGB sensors fail.

Despite the advantages of privacy protection and illumination robustness, current event-based emotion analysis methods \cite{adra2024beyond, becattini2024neuromorphic, mastropasqua2025exploring} perform poorly under such complex settings. We identify three key reasons: First, existing event-based datasets are limited in scale, struggling to support the training of deep learning models. Second, environmental setups are simple, and do not take into account for real-world variables such as diverse lighting, varying distances, and occlusion. Third,  there is a lack of exploration into multimodal mechanisms. Human emotion is a multidimensional expression comprising visual, acoustic, and linguistic cues. Relying solely on a single visual modality is insufficient to represent high-level semantics and capture the complete generative logic of emotions. Thus, a dedicated benchmark for these settings, together with a multimodal emotion analysis model, is needed to advance research.

\begin{figure}
    \centering
  \includegraphics[scale=1.1]{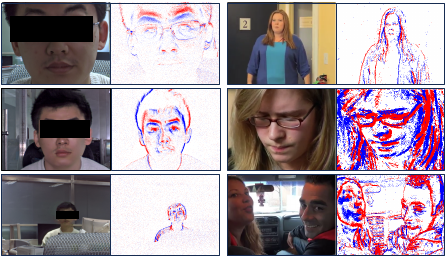}
    \caption{ Examples of Emo-DVS. The left columns show real-world captured samples, the right shows simulated samples (based on data from VideoEmotion8, Ekman6, and etc.)
    } 
    \label{fig:dataset}
   \vspace{-3mm}
\end{figure}

To this end, we construct Emo-DVS and establish a benchmark. First, we introduce Emo-DVS,  as shown in Fig. ~\ref{fig:dataset}, a large-scale event-based emotion analysis dataset with varying dynamic illumination, which combines real-world recordings with fine-grained FAU annotations to capture low-level physical dynamics, and simulated data for high-level emotional semantics. Second, based on the proposed dataset, we establish a challenging multimodal emotion analysis benchmark. Unlike previous event-based studies limited to a single visual modality, this benchmark defines a tri-modal emotion classification task comprising event, audio, and text, aiming to explore the cross-modal complementary mechanisms between bio-inspired visual, acoustic and linguistic semantics. Finally, to provide a strong baseline, we propose an Information-Guided Gated Fusion (IGF) framework. By first pre-training an event encoder on the FAU subset of Emo-DVS, the framework effectively captures fine-grained low-level physical dynamics. The extracted spatio-temporal priors, combined with acoustic and textual features, are then passed through an adaptive modality gating module to filter specific noises. Subsequently, it uses mutual information maximization to align cross-modal representations for emotion analysis.

Extensive experiments on Emo-DVS demonstrate that the proposed framework effectively leverages cross-modal cues, establishing a state-of-the-art performance for privacy-preserving emotion analysis across varying illumination conditions. Our main contributions are summarized as follows:
\vspace{-1mm}
\begin{itemize}
\item We construct Emo-DVS, the large-scale event-based emotion analysis dataset that couples dynamic illumination conditions with hierarchical FAU.
\item We propose a challenging multimodal emotion classification benchmark capable of exploring the complementary mechanisms among  event streams, text, and audio.
\item We introduce an adaptive modality gating scheme and a mutual information maximization strategy, forming the robust IGF framework for cross-modal emotion analysis.
\end{itemize}

\section{Related Work}

\subsection{Event-based emotion analysis datasets}



Event-based emotion analysis has attracted increasing attention, but public datasets remain scarce. As summarized in Table~\ref{table2:Emo-DVS}, NEFER \cite{berlincioni2023neuromorphic}, FACEMORPHIC \cite{becattini2024neuromorphic}, Emo-SNN \cite{mastropasqua2025exploring}, and ERD \cite{becattini2022understanding} provide paired RGB and event data with different annotations. VETEX \cite{adra2024beyond} further includes RGB, thermal, and event modalities with Action Unit annotations. Other studies \cite{verschae2023event, berlincioni2024neuromorphic} simulate event streams from RGB datasets such as CK+ \cite{lucey2010extended} and MMI \cite{pantic2005web} using v2e \cite{hu2021v2e}. However, synthetic data may suffer from domain gaps with real recordings.

Existing datasets are also limited in scale, annotation consistency, and environmental diversity. To address these issues, we introduce Emo-DVS, a large-scale event-based emotion dataset featuring diverse recording tasks, lighting conditions, and FAU annotations. It provides a challenging benchmark for emotion recognition in complex environments.
\newcommand{\cmark}{\raisebox{-1.2pt}{\CheckmarkBold}}
\newcommand{\xmark}{\raisebox{-0.8pt}{\XSolidBrush}}

\begin{table*}[]
\centering
\renewcommand{\arraystretch}{1.2}
\setlength{\tabcolsep}{8pt}
\resizebox{\textwidth}{!}{
\begin{tabular}{cccccccccccccc}
\toprule
\multirow{2}{*}{\textbf{Year}} & \multirow{2}{*}{\textbf{Dataset}} & \multirow{2}{*}{\textbf{Scale}} & \multirow{2}{*}{\textbf{Class}} & \multirow{2}{*}{\textbf{Users}} & \multirow{2}{*}{\textbf{Res.*}}       & \multirow{2}{*}{\textbf{Synch.}}   & \multirow{2}{*}{\textbf{Real}}         & \multirow{2}{*}{\textbf{Occ.}}         & \multirow{2}{*}{\textbf{Light*}}       & \multicolumn{4}{c}{\textbf{Annotations}}                                                                                                                               \\ \cline{11-14} 
                      &                          &                        &                        &                        &                               &                                &                               &                               &                               & \textbf{LMs}            & \textbf{BBox}           & \textbf{Emo.}           & \textbf{AUs}            \\ \hline
2023                  & e-CK+  \cite{lucey2010extended}                  & 327                    & 7 & 93                     & $346 \times 260$                        & \xmark                & \raisebox{-0.4ex}{\XSolidBrush} & \xmark                        & \xmark                        & \cmark         & \xmark         & \cmark         & \cmark         \\ 
2023                  & e-MMI \cite{lucey2010extended}                   & 2,000 & 9 & 75                     &       $346 \times 260$                 & \xmark                & \xmark                        & \xmark                        & \xmark                        & \cmark         & \xmark         & \cmark         & \cmark         \\ 
2022                  & ERD     \cite{becattini2022understanding}                  & 455                    & 3 & 25                     &   $640 \times 480$                      & \cmark               & \cmark                        & \xmark                        & \xmark                        & \xmark         & \cmark         & \xmark         & \xmark         \\ 
2023                  & NEFER  \cite{berlincioni2023neuromorphic}                  & 609                    & 8                      & 29                     &  $1280 \times 720$                       & \cmark             & \cmark                        & \xmark                        & \xmark                        & \cmark         & \cmark         & \cmark         & \xmark         \\ 
2024                  & FACEMORPHIC  \cite{becattini2024neuromorphic}             & 3,148                   & 24                     & 64                     &   $1280 \times 720$                      & \cmark              & \cmark                        & \xmark                        & \xmark                        & \cmark         & \cmark         & \xmark         & \cmark         \\ 
2024                  & VETEX \cite{adra2024beyond}                    & 2,506                   & 7                      & 20                     &       $346 \times 260$                 & \cmark                & \cmark                        & \cmark                        & \cmark                        & \xmark         & \xmark         & \xmark         & \cmark         \\ 
2025                  & Emo-SNN\cite{mastropasqua2025exploring}                 & 510                    & 21                     & 7                      &     $1280 \times 720$                     & \cmark              & \cmark                        & \xmark                        & \cmark                        & \xmark         & \xmark         & \xmark         & \cmark         \\ \hline
2026                  & Ours (Emo-DVS)             & 13,066                  & 29+8*                     & 15+2,721*                     &  $1280 \times 720$                        & \cmark              & \cmark                        & \cmark                        & \cmark                        & \cmark         & \cmark         & \cmark         & \cmark         \\ \bottomrule
\end{tabular}
}
\caption{Comparison of datasets for event-based emotion analysis. Res., Synch., Occ., LMs, BBox, Emo., and AUs denote resolution, synchronization, occlusion, landmarks, bounding box, emotion, and facial action units. Here, Res.* denotes the resolution of event camera, Light* indicates whether simulated natural light is present, 29+8* in Class indicates 29 action units (AUs) and 8 emotion categories, and 15+2,721* in Users denotes 15 participants and 2,721 personalities from public datasets.}
\label{table2:Emo-DVS}

\vspace{-15pt}
\end{table*}

\subsection{Video Emotion Analysis}


Video Emotion Analysis (VEA) is a task in affective computing, with applications in psychological analysis \cite{cai2022eeg, jia2022s, qian2024controllable, zhang2024extdm}, human--computer interaction \cite{li2024learning, liu2023pgfnet, liu2023progressive}, and intelligent surveillance \cite{liu2023progressive, huang2024alignsam, zhao2024lake}. Emotion models are categorized into categorical and dimensional approaches. Categorical methods \cite{wang2022ease, mittal2020m3er, sikka2013multiple} represent emotions as discrete classes, such as Ekman's six basic emotions \cite{ekman1999basic}, whereas dimensional methods \cite{schlosberg1954three} describe them in a continuous space, typically using Valence, Arousal, and Dominance.

Deep-learning methods increasingly adopt multimodal emotion analysis \cite{hu2024robust, hazarika2018icon, poria2019emotion, wei2023multi, zheng2023facial, hu2022mm}, but most combine RGB, audio, and text. Tri-modal emotion analysis using neuromorphic event data remains underexplored. We address this gap by introducing a benchmark that integrates event streams, audio, and text.

\vspace{-2mm}

\section{The Emo-DVS Dataset}
\subsection{Setting and Protocol}

\begin{figure*}
    \centering
  \includegraphics[scale=1.55]{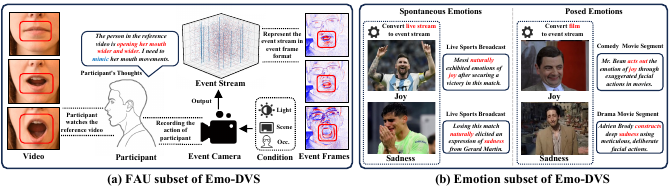}
    \caption{ The construction of the Emo-DVS dataset. (a) Data collection for the FAU subset. Participants watch demonstration videos and internalize the target action. Following this cognitive mapping, event cameras record the entire imitation process. The sensors output raw event streams, which then were converted into event frames. (b) In the spontaneous paradigm, we select unscripted live streams to ensure the induction of the characters' emotional states by the external environment. In the posed paradigm, we select videos dominated by performance, where the expressions of the selected characters are performed based on experience after receiving instructions. Finally, we convert these selected videos to event streams.
    } 
    \label{fig3:dataset}
   \vspace{-3mm}
\end{figure*}

To bridge the gap between laboratory-constrained datasets and real-world scenarios, we construct Emo-DVS, a large-scale multimodal event-based benchmark for emotion analysis. The dataset contains 13,066 video clips in total, including 4,042 clips in the \textbf{FAU set} and 9,024 clips in the \textbf{Emotion set}. As shown in Fig.~\ref{fig3:dataset}, Emo-DVS follows a hierarchical design that couples low-level facial dynamics with high-level emotional semantics. The \textbf{FAU set} forms the foundational physical layer by focusing on fine-grained facial action units (FAUs), while the \textbf{Emotion set} captures high-level emotional semantics under spontaneous and posed expression paradigms.

Collecting natural emotional expressions through real-world recordings is challenging, since the awareness of being recorded often introduces psychological cues and self-conscious behavioral bias, which may compromise expression authenticity. To alleviate this issue, we construct the Emotion set by curating large-scale web videos and organizing them into \textit{spontaneous} and \textit{posed} paradigms. This design captures more diverse and natural emotional behaviors while reducing recording-induced bias. In addition, the real-world captured event portion of Emo-DVS already provides a solid real-event foundation for the benchmark.

\noindent\textbf{The FAU set of Emo-DVS.} To enrich open-source resources for event-based emotion analysis, we construct a video subset in which participants perform specific Action Units (AUs) defined by the Facial Action Coding System (FACS). This subset contains 29 distinct AUs, including 21 facial muscle activations and 8 macro actions involving head movements. Data acquisition follows a guided elicitation protocol. For each AU, participants first watch a demonstration video and then imitate the target action at a natural speed. Therefore, these labels  indicate elicited target actions rather than expert-certified FACS annotations.

We employ the DVSync hybrid camera system as the capture framework to ensure high-fidelity multimodal alignment. This sensing platform integrates a Prophesee DVS event sensor at $1280\times720$ resolution with a synchronized IMX415 RGB sensor at $3840\times2160$ resolution within a unified optical architecture. The hardware-level synchronization resolves the spatial-temporal misalignment commonly encountered in neuromorphic data collection. During acquisition, the DVSync system is mounted on a fixed tripod and recordings are conducted in diverse environments, including bedrooms, kitchens, and classrooms to improve environmental diversity and practical relevance. Unlike previous event-based datasets that focus only on cropped facial regions, we adopt a multi-distance recording protocol, including near-distance face-centered views, medium-distance upper-body views, and far-distance full-body views while preserving facial visibility. Moreover, to exploit the high dynamic range of event cameras, data collection is performed under illumination levels ranging from 100--150 lux down to extreme low-light conditions of 2--5 lux. Under such minimal illumination, where conventional RGB sensors often fail, the event camera retains the ability to capture fine-grained facial dynamics. To further improve robustness against facial occlusion, participants are recorded both with and without glasses.

\noindent\textbf{The Emotion set of Emo-DVS.} Corresponding to the FAU Set, the Emotion Set captures higher-level semantics and contains 9,024 video sequences. These sequences cover diverse categorical emotions under two distinct emotion-generation mechanisms, namely posed and spontaneous. For the posed setting, we collect high-quality acted performances, including cinematic clips processed with event simulation methods such as v2e, to capture intention-driven affective expressions. In contrast, the spontaneous setting consists of unconstrained real-world recordings in which emotions are naturally elicited by external stimuli. All sequences are annotated with reliable semantic labels through cross-verification by independent annotators.

The spontaneous subset contains 3,582 sequences collected from platforms such as TikTok, YouTube, and Instagram, with each source contributing about 600 samples. We select unscripted videos from stimulus-driven scenarios, including sports events, live concerts, and lecture halls, covering six emotion categories defined by Ekman’s criteria \cite{ekman1999basic}. The duration of each sequence is limited to 10--15 seconds, which matches the natural evolution of emotional expression from onset to peak and offset while avoiding static frames with limited neuromorphic information. The posed subset contains 5,442 sequences. Instead of using unscripted recordings, it is built from intention-driven performances mainly derived from the VideoEmotion8 \cite{jiang2014predicting} and Ekman6 \cite{xu2016heterogeneous} benchmarks. These sequences represent scenarios in which facial dynamics are closely associated with goal-directed semantic expressions.

After collection and cropping, all RGB videos are converted into event streams using the v2e \cite{hu2021v2e} simulator. All generated event data are standardized and archived in HDF5 format, ensuring temporal precision and enabling efficient data loading for model training.

\subsection{Annotation Strategy}

\textbf{Text and Audio Modalities Annotation.} We developed different feature extraction methods for each subset, to support multimodal emotion analysis. Within the FAU subset, we built a fine-grained semantic mapping library alongside a multi-template strategy. This approach converts discrete action labels into structured natural language descriptions.For example, we transform the specific label FAU1 into the descriptive text ``\textit{a face showing an inner brow raiser}''. This conversion resolves the cross-modal feature alignment limitations in discrete labels. Conversely, for the Emotion set featuring naturalistic contexts, we used FFmpeg to extract and standardize the audio tracks from the raw videos. Subsequently, applying Qwen3-ASR \cite{shi2026qwen3} which is an automatic speech recognition model transcribes these tracks into text dialogues. These aligned audio signals and semantic texts provide vital contextual supplements for multimodal emotion analysis in complex scenarios.

\noindent\textbf{Bounding Boxes and Facial Landmarks.} The lack of absolute texture and intensity information renders precise spatial localization directly on asynchronous event streams a challenge. Using the hardware-level spatial-temporal alignment of the recording setup alongside the exact frame-to-event correspondence from the v2e simulations, we applied face detection algorithm \cite{kartynnik2019real} and facial landmark estimator \cite{yu2024yolo} directly to the aligned RGB video frames, thereby extracting accurate bounding boxes and dense facial landmarks. Finally, subjecting all auto-generated spatial annotations to rigorous frame-by-frame visual validation by domain experts guarantees absolute accuracy and temporal consistency.



\noindent\textbf{Emotion Annotation.} Emotion annotation aims to describe the emotional states of individuals in the videos. For the FAU subset, categorical labels were derived using a standard Facial Action Unit mapping table. For the posed paradigm of the Emotion Set, we directly inherited validated annotations from the source benchmarks. For the spontaneous paradigm, initial emotion labels were obtained from original video tags and contextual cues. Multiple trained annotators then conducted independent cross-checking and voting, and the final labels for the eight emotion categories were determined by majority agreement to ensure annotation accuracy and objectivity.

\noindent\textbf{Facial Action Units Annotation.} Facial Action Unit annotations provide fine-grained descriptions of facial muscle activations. In the FAU subset, video-level labels correspond to the target actions demonstrated to participants. However, posed expressions may also trigger unintended non-target action units. We therefore employed OpenFace \cite{baltrusaitis2018openface} to identify the temporal boundaries of these activations, followed by manual verification by domain experts to ensure annotation accuracy and reliability.

\subsection{Dataset Statistics and Evaluation Protocol}

\noindent\textbf{Dataset Statistics.} The FAU subset contains 4,042 clips covering 29 distinct AUs, with approximately 140 samples per AU to reduce long-tail effects. It includes 15 participants aged 25 to 56 years, with a male-to-female ratio of approximately 2:1. The Emotion subset contains 9,024 sequences, including 3,582 spontaneous clips spanning six emotion categories and 5,442 posed clips.

\noindent\textbf{Evaluation Protocol.} Given the identity-sensitive nature of the FAU subset, we adopt a cross-subject splitting strategy to prevent identity leakage and ensure a reliable  generalization evaluation. Specifically, participants are divided into mutually exclusive training, validation, and test sets with a ratio of 70\%, 10\%, and 20\%, respectively. Under this protocol, identities in the test set remain unseen throughout model training. For the Emotion subset, we adopt a 70\%/10\%/20\% split for training, validation, and testing, respectively. Since this subset is collected from large-scale heterogeneous web videos and benchmark sources, most sequences correspond to different individuals and distinct scenarios. To reduce potential data leakage, all clips derived from the same original source video are assigned to the same split, preventing near-duplicate content from appearing across training and testing. This source-disjoint protocol helps mitigate identity overlap and cross-source leakage, thereby supporting fair evaluation and reproducibility.

\begin{figure*}
    \centering
  \includegraphics[scale=1]{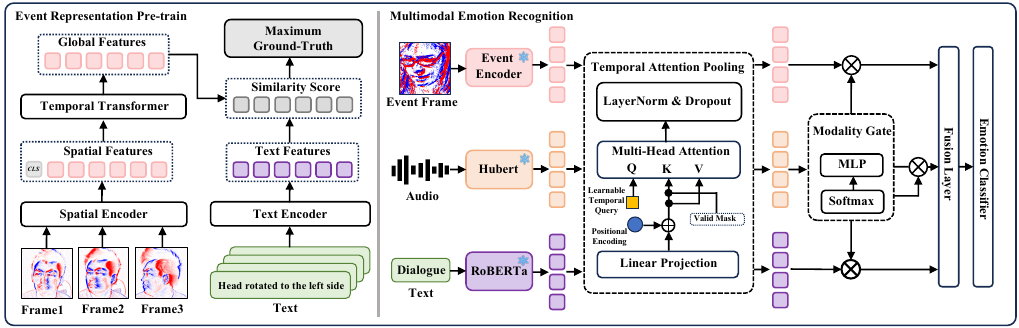}
    \caption{ Overview of our framework. In the pre-training stage, the FAU subset is used for pre-training. Event frames are first processed by a Spatial Encoder to extract frame-wise spatial features. These features are then combined with a temporal CLS token and fed into a Temporal Transformer to obtain global representations, which are aligned with text features using a contrastive loss or similarity-based objective. In the emotion recognition stage, features from the three modalities are extracted by their respective encoders and aggregated by Temporal Attention Pooling (TAP) to obtain compact modality-specific features. Modality gate then applied to regulate each modality, and the resulting features are finally fused for emotion classification.
    } 
    \label{fig4:method}
   \vspace{-3mm}
\end{figure*}

\subsection{Privacy and Ethical Considerations}

To protect participant privacy and ensure ethical research conduct, this study has obtained formal ethical approval from our institutional research ethics committee. All participants in real-world data collection have provided written informed consent for non-commercial academic usage, with the right to withdraw at any time. For third-party public video data, we strictly follow academic research reasonable use specifications and only include content that is publicly accessible and suitable for research purposes. 
The event-based modality inherently suppresses static facial textures and identity details, providing a low-identity-exposure representation. Access is restricted to verified academic researchers under a formal Data Use Agreement (DUA), which strictly prohibits commercial usage, unauthorized distribution, and any attempts to re-identify participants. Detailed ethical protocols are included in the supplementary materials.

To ensure full reproducibility and facilitate future research in this field, we will make our complete codebase publicly available and release the Emo-DVS dataset to verified academic researchers under a formal data use agreement.


\section{Method}

\subsection{Preliminaries}

\textbf{Event Camera.} Unlike standard frame-based cameras, event cameras \cite{yang2024event, pan2020high, pan2020single, Yang_2025_CVPR} asynchronously record per-pixel illumination changes rather than absolute intensity. Specifically, an event $e_i = (x_i, y_i, t_i, p_i)$ is triggered when the logarithmic intensity change at coordinate $(x_i, y_i)$ exceeds a predefined contrast threshold $C > 0$,
\begin{equation}\log I(x_i, y_i, t_i) - \log I(x_i, y_i, t_i - \Delta t) \ge p_i \cdot C,\label{eq:event_generation}
\end{equation}
where $\Delta t$ is the time elapsed since the last event at $(x_i, y_i)$, and $p_i \in \{+1, -1\}$ denotes the polarity representing brightness changes. Consequently, the camera outputs a highly sparse, asynchronous event stream $\mathcal{E} = \{e_i\}_{i=1}^{N}$. Although this paradigm provides exceptional temporal resolution and high dynamic range \cite{Gallego_2018_CVPR}, the sparse nature of $\mathcal{E}$ is inherently incompatible with standard dense neural networks. To bridge this gap, we aggregate events within a time window $\Delta T$ into a dense 3D event frame $F$. The pixel value at $(x, y)$ is computed by accumulating event polarities,
\begin{equation}F(x, y) = \sum_{e_i \in \mathcal{E}_{\Delta T}} p_i \delta(x - x_i, y - y_i),\end{equation}
where $\mathcal{E}_{\Delta T}$ is the event subset within $\Delta T$, and $\delta(\cdot, \cdot)$ is the Kronecker delta function. This dense representation is then directly processed by subsequent spatial-temporal architectures.

\subsection{Event Representation Pre-training}

The event stream is sparse and lacks RGB texture, making it susceptible to background noise. To extract robust facial dynamic priors, we pre-train our event spatial-temporal encoder on a  FAU recognition task, using the FAU subset of the Emo-DVS dataset.

As shown in Fig. \ref{fig4:method}, we first represent the raw event stream as a sequence of event frames $\mathbf{V} \in \mathbb{R}^{T \times C \times H \times W}$ by aggregating events into $T$ temporal bins. Given event frames $\mathbf{V}$, we employ a frozen pre-trained CLIP spatial backbone $\mathcal{F}_{s}$ to extract frame-level representations $\mathbf{X}_s = \mathcal{F}_{s}(\mathbf{V}) \in \mathbb{R}^{T \times d}$. To accommodate the non-square input, the pre-trained spatial positional embeddings are interpolated.

To capture macro-level temporal deformations, we introduce a lightweight Temporal Transformer $\mathcal{F}_{t}$. We prepend a learnable temporal class token $x_{\text{cls}} \in \mathbb{R}^{d}$ to $\mathbf{X}_s$ and inject temporal positional embeddings $\mathbf{P} \in \mathbb{R}^{(T+1) \times d}$. The global video representation $\mathbf{v}_e \in \mathbb{R}^d$ is derived from the first output token,
\begin{equation}
\mathbf{v}_e = \mathcal{F}_{t}([x_{\text{cls}} ; \mathbf{X}_s] + \mathbf{P})[0],
\end{equation}
where $[\cdot ; \cdot]$ denotes the concatenation operation. 

To bridge the modality gap and enhance semantic richness, we optimize a text-anchored contrastive objective instead of using a standard linear classifier. Using the frozen CLIP text encoder $\mathcal{F}_{\text{text}}$, we generate textual anchors $\mathbf{t}_c = \mathcal{F}_{\text{text}}(\mathbf{p}_c) \in \mathbb{R}^d$ for each FAU category $c$. Here, $\mathbf{p}_c$ integrates the class name with continuous learnable context tokens. The pre-training loss is formulated as,
\begin{equation}
\mathcal{L}_{\text{FAU}} = \mathcal{L}_{\text{CE}}(\sigma \langle \bar{\mathbf{v}}_e, \bar{\mathbf{t}}_c \rangle, y),
\end{equation}where $\mathcal{L}_{\text{CE}}$ is the cross-entropy loss, $\sigma$ is a learnable temperature scalar, $y$ is the ground-truth label, and $\langle \bar{\mathbf{v}}_e, \bar{\mathbf{t}}_c \rangle$ denotes the cosine similarity between the $L_2$-normalized representations. This vision-language alignment forces the encoder to capture precise dynamic cues while resisting modality-irrelevant noise.

\subsection{Information-Guided Multimodal Fusion}

To achieve robust tri-modal emotion analysis on the emotion subset of the Emo-DVS, we design an Information-Guided Multimodal Fusion (IGF) framework which is detailed as follows.

\noindent\textbf{Temporal Attention Pooling.} We introduce a Temporal Attention Pooling ($\mathcal{T}$) mechanism to distill variable-length sequences from event-based visual frames $\mathbf{X}_v$ encoded by a pre-trained event encoder, alongside acoustic features $\mathbf{X}_a$ from HuBERT \cite{hsu2021hubert} linguistic features $\mathbf{X}_t$ from RoBERTa \cite{liu2019roberta}. For each modality $m \in \{v, a, t\}$, the raw sequence is projected into a latent space $\mathbb{R}^d$ and augmented with a sinusoidal positional encoding $\mathbf{P} \in \mathbb{R}^{T \times d}$. We define a learnable temporal query $\mathbf{q} \in \mathbb{R}^{1 \times d}$ to aggregate the sequence through a multi-head attention  $\mathcal{A}(\cdot)$ followed by a layer normalization $\mathcal{N}(\cdot)$,

\begin{equation}\mathbf{h}_m = \mathcal{N}\left( \mathcal{A}(\mathbf{q}, \tilde{\mathbf{X}}_m, \tilde{\mathbf{X}}_m) \right),
\end{equation}where $\tilde{\mathbf{X}}_m$ denotes the position-aware features. This process yields a compact unimodal descriptor $\mathbf{h}_m \in \mathbb{R}^d$ for each modality by adaptively attending to the most informative temporal segments.

\noindent\textbf{Dynamic Modality Gating.} To dynamically weigh the contribution of each modality based on the current input context, we design a modality gate. Initially, the pooled unimodal features are concatenated along the channel dimension and fed into a non-linear mapping network $\mathcal{F}_{gate}(\cdot)$, followed by a normalization function $\sigma(\cdot)$ to compute the adaptive weights for each modality,

\begin{equation}\mathbf{g} = \sigma\left( \mathcal{F}_{gate}\left( [\mathbf{h}_v; \mathbf{h}_a; \mathbf{h}_t] \right) \right),\end{equation}
where $\mathbf{g} = [g_v, g_a, g_t] \in \mathbb{R}^3$ denotes the modality contribution vector, $\mathcal{F}_{gate}(\cdot)$ is an MLP layer, $\sigma(\cdot)$ represents the Softmax activation function. Subsequently, the raw unimodal features are scaled by their corresponding weight scalars to obtain the modulated representations, $\mathbf{h}^*_m = g_m \cdot \mathbf{h}_m$. This mechanism ensures that the network adaptively focuses on the most discriminative modalities while effectively suppressing the interference of redundant noise.

\begin{table*}[]
\centering
\renewcommand{\arraystretch}{1.2}
\setlength{\tabcolsep}{8pt}
\resizebox{\textwidth}{!}{
\begin{tabular}{cccccccccccc}
\toprule
\multirow{2}{*}{\textbf{Type}} & \multirow{2}{*}{\textbf{Method}} & \multirow{2}{*}{\textbf{Pubish}} & \multicolumn{3}{c}{\textbf{FAU subset}}          & \multicolumn{3}{c}{\textbf{Ekman6*}}             & \multicolumn{3}{c}{\textbf{VideoEmotion8*}}      \\ \cline{4-12} 
                               &                                  &                                  & Acc            & mAcc           & F1             & Acc            & mAcc           & F1             & Acc            & mAcc           & F1             \\ \hline
\multirow{6}{*}{RGB $\to$ Event}   & MoCov3 \cite{chen2021empirical}                          & ICCV21                           & 27.48          & 27.39          & 28.51          & 32.77          & 32.77          & 32.56          & 28.15          & 27.42          & 30.04          \\
                               & VAANet  \cite{Zhao2020AnEV}                          & AAAI20                           & 30.34          & 29.94          & 29.26          & 35.29          & 35.14          & 35.20          & 31.49          & 30.33          & 31.25          \\
                               & DFAN  \cite{9102808}                           & ICME20                           & 32.15          & 32.76          & 32.30          & 39.11          & 39.05          & 38.38          & 34.15          & 32.54          & 35.02          \\
                               & MGMAE   \cite{huang2023mgmae}                         & ICCV23                           & 31.01          & 30.21          & 29.88          & 38.03          & 37.15          & 38.40          & 34.11          & 33.09          & 31.48          \\
                               & MART   \cite{zhang2024mart}                           & CVPR24                           &  \underline{36.76}    & \underline{35.73}    & \underline{37.21}    & \underline{40.99}    & \underline{39.83}    & \underline{40.24}    & \underline{35.33}    & \underline{34.21}    & \underline{35.93}    \\
                               & MILAN    \cite{hou2025masked}                        & ICASSP25                         & 33.49          & 33.72          & 34.01          & 38.21          & 37.57          & 37.62          & 34.45          & 33.12          & 30.52          \\ \hline
\multirow{6}{*}{Event $\to$ Event} & ERD  \cite{becattini2022understanding}                            & TII22                            & 20.35          & 21.77          & 21.83          & 19.01          & 15.53          & 16.26          & 23.85          & 12.50          & 9.19           \\
                               & NEFER  \cite{berlincioni2023neuromorphic}                           & CVPRW23                          & 29.16          & 28.81          & 28.04          & 28.44          & 27.52          & 28.69          & 33.95          & 24.52          & 27.66          \\
                               & FACEMORPHIC   \cite{becattini2024neuromorphic}                     & ECCV24                           & 31.50          & 29.98          & 30.14         & 37.59          & 29.40          & 37.28          & 33.49          & 25.50          & 29.82          \\
                               & VETEX    \cite{adra2024beyond}                         & ICPR24                           & 28.96          & 27.61          & 27.45         & 28.20          & 16.67          & 12.40          & 24.73          & 12.50          & 9.80           \\
                               & Emo-SNN    \cite{mastropasqua2025exploring}                       & ICCVW25                          & 30.47          & 22.09          & 28.27          & 38.89          & 19.04          & 29.13          & 33.94          & 27.62          & 30.62          \\ \cline{2-12} 
                               & Ours                             &  -                                & \textbf{41.05} & \textbf{40.38} & \textbf{40.44} & \textbf{56.40} & \textbf{55.96} & \textbf{55.85} & \textbf{42.54} & \textbf{41.51} & \textbf{42.99} \\ \bottomrule
\end{tabular}}
\caption{ Comparison of emotion classification results on the Ekman6* and VideoEmotion8* in emotion subsets and FAU subset. The best result for each metric is in bold, and the second-best is underlined. Here, RGB $\to$  Event denotes using a
RGB-based method on the event dataset, and Event $\to$  Event is defined analogously. Tab. 3 also follows the same conventions.}
\label{table2:result1}
\vspace{-15pt}

\end{table*}

\begin{table}[]
\centering
\renewcommand{\arraystretch}{1.2}
\setlength{\tabcolsep}{14pt}
\resizebox{\linewidth}{!}{
\begin{tabular}{cccc}
\toprule
\multirow{2}{*}{\textbf{Type}} & \multirow{2}{*}{\textbf{Method}} & \multicolumn{2}{c}{\textbf{Ekman6}} \\ \cline{3-4} 
                               &                                  & Acc              & F1              \\ \hline
\multirow{6}{*}{RGB $\to$ RGB}      & MoCov3 \cite{chen2021empirical}                          & 47.47            & 46.81           \\
                               & VAANet    \cite{Zhao2020AnEV}                         & 50.24            & 49.54           \\
                               & DFAN  \cite{9102808}                           & 54.30           & 52.17           \\
                               & MGMAE   \cite{huang2023mgmae}                         & 51.98            & 52.36           \\
                               & MART   \cite{zhang2024mart}                          & 52.66            & 52.64           \\
                               & MILAN    \cite{hou2025masked}                        & 50.80            & 50.40           \\ \hline
Event $\to$ RGB                      & Ours                             & \textbf{59.43}            & \textbf{57.12}           \\ \hline
\end{tabular}
}
\caption{Results of  emotion classification on Ekman6.}
\label{table3:result2}
\vspace{-20pt}
\end{table}

\noindent\textbf{Information-Guided Multimodal Fusion.} Rather than simply concatenating the features, we construct a comprehensive fusion space that captures both modality-specific and shared semantics. First, the gated features $\mathbf{h}^*_m$ are projected into a common fusion space via modality-specific mappings to obtain $\mathbf{f}_m$. A shared representation $\mathbf{f}_{sh}$ is derived by projecting the sum of the gated features: $\mathbf{f}_{sh} = \mathcal{F}_{sh}(\mathbf{h}^*_m )$. Furthermore, the gate values themselves are projected to form a contextual gate feature $\mathbf{f}_{gate}$. The final composite representation $\mathbf{z}_{c}$ is formed by concatenating these components, 
\begin{equation}
\mathbf{z}_{c} = \left[\mathbf{f}_m, \mathbf{f}_{sh}, \mathbf{f}_{gate} \right].
\end{equation}

The vector ${z}_{c}$ is then processed by an MLP layer $\mathcal{F}_{fus}(\cdot)$ to generate the final fused representation $\mathbf{z}_{fuse}$, which serves as input to the classification head to predict the emotions.

\noindent\textbf{Optimization.} The entire framework is optimized end-to-end via a joint multi-task objective comprising a primary classification loss and three auxiliary constraints. First, to encourage modality-invariant semantics, we compute a Mutual Information (MI) maximization loss $\mathcal{L}_{mi}$, defined as the average of symmetric InfoNCE losses across all modality pairs,

\begin{equation}
\mathcal{L}_{mi}(\mathbf{z}^{mi}_i, \mathbf{z}^{mi}_j) = \frac{1}{2} \left( \ell(\mathbf{z}^{mi}_i, \mathbf{z}^{mi}_j) + \ell(\mathbf{z}^{mi}_j, \mathbf{z}^{mi}_i) \right),
\end{equation} where $\mathbf{z}^{mi}_m$ represents the gated features $\mathbf{h}^*_m$ projected into a contrastive space, and $\ell(\cdot, \cdot)$ is the cross-entropy over temperature-scaled cosine similarity logits. Then, we incorporate an alignment loss $\mathcal{L}_{alg}$ to align the final fused feature with the shared auxiliary representation, and a redundancy-minimization loss $\mathcal{L}_{red}$ to orthogonalize the modality-specific residuals $\mathbf{r}_m$,

\begin{equation}\mathcal{L}_{alg} = 1 - \rho(\mathbf{z}_{fuse}, \mathbf{f}_{sh}), \quad \mathcal{L}_{red} = \sum_{m \neq n} \rho^2(\mathbf{r}_m, \mathbf{r}_n),\end{equation} where $\mathbf{r}_m = \mathbf{h}_m - \frac{1}{3} \sum_{k \in \{v,a,t\}} \mathbf{h}_k$ and $\rho(\cdot)$ denotes cosine similarity. The optimization is the weighted sum of these components,

\begin{equation}\mathcal{L} = \mathcal{L}_{ce} + \lambda_{mi} \mathcal{L}_{mi} + \lambda_{alg} \mathcal{L}_{alg} + \lambda_{red} \mathcal{L}_{red},\end{equation} where  $\lambda_{(\cdot)}$ are the corresponding balancing coefficients.

\section{Experiment}

\subsection{Experimental Setup}

\textbf{Datasets and Baseline Models}. We first pre-train our model on the FAU subset of Emo-DVS, and then conduct downstream emotion classification on the VideoEmotion8* and Ekman6* subsets of the Emotion subset and FAU subset. As shown in Tab.~\ref{table2:result1}, we compare our method with representative approaches from multiple perspectives. Since existing event-based methods are mainly designed for unimodal event streams and there are currently no event-based tri-modal baselines specifically designed for emotion classification, we include several representative event-based methods, including \cite{becattini2022understanding}, \cite{berlincioni2023neuromorphic}, \cite{becattini2024neuromorphic}, \cite{adra2024beyond}, and \cite{mastropasqua2025exploring}. We further compare with RGB-based multimodal methods, including \cite{Zhao2020AnEV}, \cite{9102808}, and \cite{zhang2024mart}, as well as video-based emotion classification methods such as \cite{chen2021empirical}, \cite{huang2023mgmae}, and \cite{hou2025masked}. To improve comparability, we evaluate all methods under a unified event-based protocol. Specifically, RGB-based and event-based methods are adapted to event-frame inputs, and the adapted models are pre-trained on the FAU subset before downstream emotion classification.

\noindent\textbf{Implementation Details}. 
For emotion recognition, we extract visual, acoustic, and linguistic features using a pre-trained event encoder, HuBERT \cite{hsu2021hubert}, and RoBERTa \cite{liu2019roberta}, respectively, with feature dimensions of 768, 1024, and 1024. Unless otherwise stated, these encoders are kept frozen during downstream training, and only the temporal aggregation, fusion, and classification modules are optimized. In our fusion network, both the hidden dimension and the fusion dimension are set to 128. A dropout rate of 0.4 is applied to the fusion backbone and projection layers. Raw event streams are aggregated into event frames using a temporal window of $\Delta T = 5000\,\mu s$. We train all models with AdamW using an initial learning rate of $2 \times 10^{-4}$, a batch size of 32, and 100 epochs. All experiments are conducted on NVIDIA RTX 4090 GPUs.

\subsection{Main Results}

Tab.~\ref{table2:result1} and Tab.~\ref{table3:result2} compare the proposed IGF framework with representative baselines on VideoEmotion8*, Ekman6* of emotion subset and FAU subset. The best results are highlighted in bold. Our main findings are summarized below (1) Under the unified evaluation protocol, our method achieves the best performance among the compared methods on both three datasets, reaching 41.05\%, 42.54\% and 56.40\% Acc, respectively. (2) Compared with RGB-based baselines, our framework shows the advantage of event-pretrained representations for modeling fine-grained facial dynamics and emotion-related temporal cues. (3) Compared with existing event-based baselines, the proposed framework yields clear performance gains, suggesting that the current benchmark benefits from structured multimodal modeling rather than unimodal event features alone. (4) As shown in Tab.~\ref{table3:result2}, we further examine the cross-modal transferability of the proposed framework on the RGB-based Ekman6 dataset. Although the model is built upon event-pretrained representations, it still achieves 59.43\% Acc and 57.12\% F1 after being transferred to the RGB emotion classification task, outperforming all compared RGB-based methods. This suggests that event pre-training does not merely learn features specific to the event domain, but instead captures facial dynamic priors that remain useful for RGB emotion recognition, providing evidence of cross-modal generalization. For brevity, we present the Ekman6 results here, while additional transfer results are provided in the supplementary material.

\begin{figure*}
    \centering
  \includegraphics[scale=0.78]{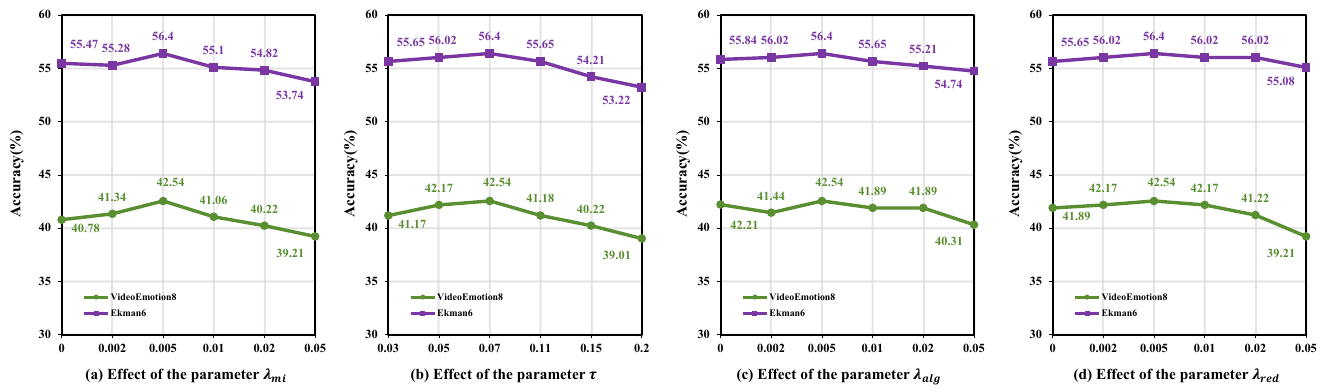}
   \vspace{-4mm}
    \caption{  Results of parameter sensitivity experiment.
    } 
    \label{fig:param}
   \vspace{-3mm}
\end{figure*}

\subsection{Ablation Studies}

\noindent\textbf{Effectiveness of Pre-training.} As shown in Tab.~\ref{table3:ablation}, removing pre-training on the FAU subset causes a substantial drop of 8.03\%. This result indicates that low-level facial dynamics learned from FAU provide an important prior for decoding sparse event streams into higher-level emotional semantics, and confirms the role of hierarchical pre-training in our framework.

\noindent\textbf{Effectiveness of Temporal Attention Pooling (TAP).} Replacing TAP with standard average pooling (w/o TAP) leads to a 1.76\% drop. This suggests that simple temporal averaging is insufficient for unconstrained emotion sequences with diverse temporal dynamics, whereas TAP benefits from a learnable global query that selectively emphasizes informative temporal segments.

\noindent\textbf{Effectiveness of Dynamic Modality Gating (MG).} We further evaluate a variant without modality gating (w/o MG), where all modalities contribute equally throughout inference. The performance degradation shows that adaptive modality weighting is beneficial, especially when different streams exhibit unequal reliability or contain modality-specific noise.

\noindent\textbf{Effectiveness of Information-Guided Multimodal Fusion.} Finally, we replace the proposed fusion module with simple feature concatenation (w/o Fusion). The inferior performance of this variant indicates that structured cross-modal fusion is more effective than naive concatenation under the current benchmark, supporting the benefit of combining gated features with information-guided constraints for robust multimodal learning.

\begin{table}[]
\centering
\renewcommand{\arraystretch}{1.2}
\setlength{\tabcolsep}{8pt}
\resizebox{\linewidth}{!}{
\begin{tabular}{cccccc}
\toprule
\textbf{Case} & \textbf{Pre-train} & \textbf{TAP} & \textbf{MG} & \textbf{Fusion} & \textbf{Results} \\
\midrule
\textit{w/o} Pre-train &  & \cmark & \cmark & \cmark & 33.24 \\
\textit{w/o} TAP  & \cmark &  & \cmark & \cmark & 39.51 \\
\textit{w/o} MG   & \cmark & \cmark &  & \cmark & 39.43 \\
\textit{w/o} Fusion   & \cmark & \cmark & \cmark &  & 40.17 \\
\midrule
\textbf{Ours}     & \cmark & \cmark & \cmark & \cmark & 42.54 \\
\bottomrule
\end{tabular}
}
\caption{Ablation study of Pre-train, TAP, MG, and Fusion on the VideoEmotion8* in Emotion subset of Emo-DVS.}
\label{table3:ablation}
\vspace{-25pt}
\end{table}

\subsection{Discussion}

\noindent\textbf{Hyperparameter and Auxiliary Loss Analysis.} As shown in Fig.~\ref{fig:param}, we analyze four key hyperparameters: $\lambda_{mi}$, $\tau$, $\lambda_{alg}$, and $\lambda_{red}$. In particular, setting $\lambda_{mi}=0$, $\lambda_{alg}=0$, or $\lambda_{red}=0$ corresponds to removing the corresponding auxiliary objective. The results show that moderate $\lambda_{mi}$ and $\tau$ achieve the best performance, while excessively large values tend to interfere with the main classification objective. By contrast, $\lambda_{alg}$ and $\lambda_{red}$ are most effective as lightweight regularizers: small but non-zero values improve consistency and complementarity, whereas overly strong constraints degrade generalization. Overall, these results suggest that the auxiliary objectives are beneficial when properly balanced.

\begin{table}[]
\centering
\renewcommand{\arraystretch}{1.2}
\setlength{\tabcolsep}{13pt}
\resizebox{\linewidth}{!}{
\begin{tabular}{ccccc}
\toprule
\multirow{2}{*}{\textbf{Modality}} & \multicolumn{2}{c}{\textbf{Ekman*}} & \multicolumn{2}{c}{\textbf{VideoEmotion8*}} \\ \cline{2-5} 
                                   & Acc               & F1              & Acc                   & F1                  \\ \hline
E                                  &45.27                   &  45.12               &   29.61                    &  26.23                   \\
A                                  &41.74                   &  42.78               &    32.12                   &   30.80                  \\
T                                  &  44.90                 &  44.33               &    33.52                   &  33.27                   \\ \hline
E+T                                &    54.17               &     53.33            &    37.99                   &   37.64                  \\
E+A                                &  47.12                 &   45.22              &     36.59                  &     36.46                \\
A+T                                &   49.35                &  49.19               &       39.39                &       39.46              \\ \hline
E+A+T(Ours)                        &    56.40               &   55.85              &      42.54                 &    42.99                 \\ \bottomrule
\end{tabular}
}
\caption{ Results of modality ablation on Ekman* and VideoEmotion8*  in Emotion subset of Emo-DVS dataset.}
\label{tab7}
\vspace{-25pt}
\end{table}

\noindent\textbf{Privacy Protection } As shown in Tab.~\ref{table6:privacy}, when evaluated with state-of-the-art face recognition method LVface \cite{you2025lvface} , RGB data suffers a high identification rate of 98.34\%, posing significant privacy risks. Conversely, the event modality decreases the face recognition accuracy by 66.22\%, which suggests that event representations suppress a large portion of identity-related appearance cues and therefore provide a more privacy-aware alternative for affective analysis. 


\noindent\textbf{Illumination Conditions.}
Tab.~\ref{table5:light} compares RGB and event modalities under different illumination conditions on the FAU subset. The event modality consistently outperforms RGB, with the large-scale gap under low light, RGB drops to 8.93\%, whereas event still reaches 37.35\%. Under medium and high light, event also remains superior, achieving 39.27\% and 38.49\%, compared with 36.28\% and 35.27\% for RGB. These results indicate that event representations are less sensitive to illumination changes and remain more stable across lighting conditions, especially when RGB appearance cues degrade.

\noindent\textbf{Analysis of Different Modality Settings.}  We further evaluate single-modality, bimodal, and trimodal settings on both Ekman* and VideoEmotion8* to analyze the contribution of each modality. As shown in Tab.~\ref{tab7}, bimodal settings consistently outperform single-modality ones on both datasets, indicating clear complementarity across modalities. On Ekman*, the best single-modality result is achieved by E with 45.27\% Acc and 45.12\% F1, while the best bimodal setting, E+T, improves the performance to 54.17\% Acc and 53.33\% F1. On VideoEmotion8*, the best single modality is T with 33.52\% Acc and 33.27\% F1, whereas the best bimodal setting, A+T, reaches 39.39\% Acc and 39.46\% F1. The full multimodal model achieves the best performance on both datasets, showing the benefit of jointly exploiting event, acoustic, and textual information.

\begin{table}[]
\centering
\renewcommand{\arraystretch}{1.2}
\setlength{\tabcolsep}{6pt}
\resizebox{\linewidth}{!}{
\begin{tabular}{ccccc}
\toprule
\textbf{Modality} & \textbf{Average} & \textbf{Low light} & \textbf{Medium light} & \textbf{High light} \\ \hline
RGB               &      27.77         &     8.93               &  36.28                    &     35.27                \\
Event             &    38.46            &     37.35               &     39.27                 &   38.49                 \\ \bottomrule
\end{tabular}
}
\caption{Comparison between RGB and Event modalities under varying illumination on FAU subset of Emo-DVS dataset.}
\label{table5:light}
\vspace{-20pt}
\end{table}

\begin{table}[]
\centering
\renewcommand{\arraystretch}{1.2}
\setlength{\tabcolsep}{11pt}
\resizebox{\linewidth}{!}{
\begin{tabular}{ccccc}
\toprule
\textbf{Method}         & \textbf{Modality} & \textbf{Acc} & \textbf{macro-acc} & \textbf{mAP} \\ \hline
\multirow{2}{*}{LVface \cite{you2025lvface}} & RGB               &       98.34         &      98.20          &  99.67            \\
                        & Event             &       32.12          &      31.28          &    33.74          \\ \bottomrule
\end{tabular}
}
\caption{Comparison of RGB and event modalities for face recognition on the FAU subset of the Emo-DVS dataset.}
\label{table6:privacy}
\vspace{-25pt}
\end{table}

\section{Conclusion}
In this work, we present Emo-DVS, a large-scale event-based emotion analysis dataset, and establish a tri-modal benchmark for robust emotion analysis in challenge illumination and privacy-constrained environments. To  align multimodal cues, we propose an Information-Guided Gated Fusion (IGF) framework that extracts spatio-temporal priors via a pre-trained event encoder and employs mutual information maximization to classify emotions. Extensive experiments demonstrate the superiority of our framework over state-of-the-art methods.

\section{Acknowledgment}

This work is supported by the National Natural Science Foundation of China (62302045), the Fundamental Research Funds for the Central Universities, and the BIT Special-Zone. 

\bibliographystyle{ACM-Reference-Format}
\balance
\bibliography{sample-base}

@inproceedings{zhao2025context,
  title={Context-Aware Academic Emotion Dataset and Benchmark},
  author={Zhao, Luming and Xuan, Jingwen and Lou, Jiamin and Yu, Yonghui and Yang, Wenwu},
  booktitle={Proceedings of the IEEE/CVF International Conference on Computer Vision},
  pages={13859--13868},
  year={2025}
}

@inproceedings{lee2025v,
  title={V-NAW: Video-based Noise-aware Adaptive Weighting for Facial Expression Recognition},
  author={Lee, JunGyu and Lee, Kunyoung and Park, Haesol and Kim, Ig-Jae and Nam, Gi Pyo},
  booktitle={Proceedings of the IEEE/CVF Conference on Computer Vision and Pattern Recognition},
  pages={5689--5696},
  year={2025}
}

@inproceedings{kim2025contextface,
  title={ContextFace: Generating Facial Expressions from Emotional Contexts},
  author={Kim, Min-jung and Kim, Minsang and Baek, Seung Jun},
  booktitle={Proceedings of the IEEE/CVF International Conference on Computer Vision},
  pages={11383--11392},
  year={2025}
}

@inproceedings{yu2025cross,
  title={Cross-Modal Facial Expression Recognition with Global Channel-Spatial Attention: Modal Enhancement and Proportional Criterion Fusion},
  author={Yu, Jun and Zheng, Yang and Wang, Lei and Wang, Yongqi and Xu, Shengfan},
  booktitle={Proceedings of the Computer Vision and Pattern Recognition Conference},
  pages={5707--5714},
  year={2025}
}

@inproceedings{wang2023rethinking,
  title={Rethinking the learning paradigm for dynamic facial expression recognition},
  author={Wang, Hanyang and Li, Bo and Wu, Shuang and Shen, Siyuan and Liu, Feng and Ding, Shouhong and Zhou, Aimin},
  booktitle={Proceedings of the IEEE/CVF conference on computer vision and pattern recognition},
  pages={17958--17968},
  year={2023}
}

@inproceedings{ma2025multimodal,
  title={Multimodal prompt alignment for facial expression recognition},
  author={Ma, Fuyan and He, Yiran and Sun, Bin and Li, Shutao},
  booktitle={Proceedings of the IEEE/CVF International Conference on Computer Vision},
  pages={12581--12591},
  year={2025}
}

@inproceedings{cabacas2025enhancing,
  title={Enhancing Facial Expression Recognition with LSTM Through Dual-Direction Attention Mixed Feature Networks and Clip},
  author={Cabacas-Maso, Josep and Ortega-Beltr{\'a}n, Elena and Benito-Altamirano, Ismael and Ventura, Carles},
  booktitle={Proceedings of the Computer Vision and Pattern Recognition Conference},
  pages={5665--5671},
  year={2025}
}

@article{chen2023multi,
  title={Multi-relations aware network for in-the-wild facial expression recognition},
  author={Chen, Dongliang and Wen, Guihua and Li, Huihui and Chen, Rui and Li, Cheng},
  journal={IEEE Transactions on Circuits and Systems for Video Technology},
  volume={33},
  number={8},
  pages={3848--3859},
  year={2023},
  publisher={IEEE}
}

@inproceedings{savchenko2025leveraging,
  title={Leveraging lightweight facial models and textual modality in audio-visual emotional understanding in-the-wild},
  author={Savchenko, Andrey and Savchenko, Lyudmila},
  booktitle={Proceedings of the Computer Vision and Pattern Recognition Conference},
  pages={5778--5788},
  year={2025}
}

@inproceedings{li2025fed,
  title={FED-PsyAU: Privacy-Preserving Micro-Expression Recognition via Psychological AU Coordination and Dynamic Facial Motion Modeling},
  author={Li, Jingting and Qian, Yu and Zhao, Lin and Wang, Su-Jing},
  booktitle={Proceedings of the IEEE/CVF International Conference on Computer Vision},
  pages={14453--14463},
  year={2025}
}

@inproceedings{zhang2023blink,
  title={In the blink of an eye: Event-based emotion recognition},
  author={Zhang, Haiwei and Zhang, Jiqing and Dong, Bo and Peers, Pieter and Wu, Wenwei and Wei, Xiaopeng and Heide, Felix and Yang, Xin},
  booktitle={ACM SIGGRAPH 2023 Conference Proceedings},
  pages={1--11},
  year={2023}
}

@article{wang2025cs3d,
  title={CS3D: An Efficient Facial Expression Recognition via Event Vision},
  author={Wang, Zhe and Song, Qijin and Peng, Yucen and Bai, Weibang},
  journal={arXiv preprint arXiv:2512.09592},
  year={2025}
}

@article{adra2025event,
  title={Event-based solutions for human-centered applications: a comprehensive review},
  author={Adra, Mira and Melcarne, Simone and Mirabet-Herranz, Nelida and Dugelay, Jean-Luc},
  journal={Frontiers in Signal Processing},
  volume={5},
  pages={1585242},
  year={2025},
  publisher={Frontiers Media SA}
}

@article{kim2025privacy,
  title={Privacy-preserving visual localization with event cameras},
  author={Kim, Junho and Kim, Young Min and Zahreddine, Ramzi and Welge, Weston A and Krishnan, Gurunandan and Ma, Sizhuo and Wang, Jian},
  journal={IEEE Transactions on Image Processing},
  year={2025},
  publisher={IEEE}
}

@article{becattini2025neuromorphic,
  title={Neuromorphic face analysis: A survey},
  author={Becattini, Federico and Berlincioni, Lorenzo and Cultrera, Luca and Del Bimbo, Alberto},
  journal={Pattern Recognition Letters},
  volume={187},
  pages={42--48},
  year={2025},
  publisher={Elsevier}
}

@inproceedings{aquib2025decoupling,
  title={Decoupling Identity Confounders for Enhanced Facial Expression Recognition: An Information-Theoretic Approach},
  author={Aquib, Mohd and Verma, Nishchal K and Akhtar, M Jaleel},
  booktitle={Proceedings of the Computer Vision and Pattern Recognition Conference},
  pages={5552--5561},
  year={2025}
}

@inproceedings{adra2024beyond,
  title={Beyond RGB: Tri-Modal Microexpression Recognition with RGB, Thermal, and Event Data},
  author={Adra, Mira and Mirabet-Herranz, Nelida and Dugelay, Jean-Luc},
  booktitle={International Conference on Pattern Recognition},
  pages={311--324},
  year={2024},
  organization={Springer}
}

@inproceedings{berlincioni2023neuromorphic,
  title={Neuromorphic event-based facial expression recognition},
  author={Berlincioni, Lorenzo and Cultrera, Luca and Albisani, Chiara and Cresti, Lisa and Leonardo, Andrea and Picchioni, Sara and Becattini, Federico and Del Bimbo, Alberto},
  booktitle={Proceedings of the IEEE/CVF Conference on Computer Vision and Pattern Recognition},
  pages={4109--4119},
  year={2023}
}

@inproceedings{becattini2024neuromorphic,
  title={Neuromorphic facial analysis with cross-modal supervision},
  author={Becattini, Federico and Cultrera, Luca and Berlincioni, Lorenzo and Ferrari, Claudio and Leonardo, Andrea and Del Bimbo, Alberto},
  booktitle={European Conference on Computer Vision},
  pages={205--223},
  year={2024},
  organization={Springer}
}

@inproceedings{mastropasqua2025exploring,
  title={Exploring spatial-temporal dynamics in event-based facial micro-expression analysis},
  author={Mastropasqua, Nicolas and Bugueno-Cordova, Ignacio and Verschae, Rodrigo and Acevedo, Daniel and Negri, Pablo and Buemi, Maria Elena},
  booktitle={Proceedings of the IEEE/CVF International Conference on Computer Vision},
  pages={4723--4732},
  year={2025}
}

@article{becattini2022understanding,
  title={Understanding human reactions looking at facial microexpressions with an event camera},
  author={Becattini, Federico and Palai, Federico and Del Bimbo, Alberto},
  journal={IEEE Transactions on Industrial Informatics},
  volume={18},
  number={12},
  pages={9112--9121},
  year={2022},
  publisher={IEEE}
}

@article{verschae2023event,
  title={Event-based gesture and facial expression recognition: A comparative analysis},
  author={Verschae, Rodrigo and Bugueno-Cordova, Ignacio},
  journal={IEEE Access},
  volume={11},
  pages={121269--121283},
  year={2023},
  publisher={IEEE}
}

@inproceedings{lucey2010extended,
  title={The extended cohn-kanade dataset (ck+): A complete dataset for action unit and emotion-specified expression},
  author={Lucey, Patrick and Cohn, Jeffrey F and Kanade, Takeo and Saragih, Jason and Ambadar, Zara and Matthews, Iain},
  booktitle={2010 ieee computer society conference on computer vision and pattern recognition-workshops},
  pages={94--101},
  year={2010},
  organization={IEEE}
}

@inproceedings{pantic2005web,
  title={Web-based database for facial expression analysis},
  author={Pantic, Maja and Valstar, Michel and Rademaker, Ron and Maat, Ludo},
  booktitle={2005 IEEE international conference on multimedia and Expo},
  pages={5--pp},
  year={2005},
  organization={IEEE}
}

@article{berlincioni2024neuromorphic,
  title={Neuromorphic valence and arousal estimation},
  author={Berlincioni, Lorenzo and Cultrera, Luca and Becattini, Federico and Bimbo, Alberto Del},
  journal={Journal of Ambient Intelligence and Humanized Computing},
  pages={1--11},
  year={2024},
  publisher={Springer}
}

@article{cai2022eeg,
  title={EEG-based emotion recognition using multiple kernel learning},
  author={Cai, Qian and Cui, Guo-Chong and Wang, Hai-Xian},
  journal={Machine Intelligence Research},
  volume={19},
  number={5},
  pages={472--484},
  year={2022},
  publisher={Springer}
}

@inproceedings{jia2022s,
  title={S 2-ver: Semi-supervised visual emotion recognition},
  author={Jia, Guoli and Yang, Jufeng},
  booktitle={European conference on computer vision},
  pages={493--509},
  year={2022},
  organization={Springer}
}

@article{qian2024controllable,
  title={Controllable augmentations for video representation learning},
  author={Qian, Rui and Lin, Weiyao and See, John and Li, Dian},
  journal={Visual Intelligence},
  volume={2},
  number={1},
  pages={1},
  year={2024},
  publisher={Springer}
}

@inproceedings{zhang2024extdm,
  title={Extdm: Distribution extrapolation diffusion model for video prediction},
  author={Zhang, Zhicheng and Hu, Junyao and Cheng, Wentao and Paudel, Danda and Yang, Jufeng},
  booktitle={Proceedings of the IEEE/CVF Conference on Computer Vision and Pattern Recognition},
  pages={19310--19320},
  year={2024}
}

@inproceedings{li2024learning,
  title={Learning background prompts to discover implicit knowledge for open vocabulary object detection},
  author={Li, Jiaming and Zhang, Jiacheng and Li, Jichang and Li, Ge and Liu, Si and Lin, Liang and Li, Guanbin},
  booktitle={Proceedings of the IEEE/CVF conference on computer vision and pattern recognition},
  pages={16678--16687},
  year={2024}
}

@article{liu2023pgfnet,
  title={PGFNet: Preference-guided filtering network for two-view correspondence learning},
  author={Liu, Xin and Xiao, Guobao and Chen, Riqing and Ma, Jiayi},
  journal={IEEE Transactions on Image Processing},
  volume={32},
  pages={1367--1378},
  year={2023},
  publisher={IEEE}
}

@inproceedings{liu2023progressive,
  title={Progressive neighbor consistency mining for correspondence pruning},
  author={Liu, Xin and Yang, Jufeng},
  booktitle={Proceedings of the IEEE/CVF conference on computer vision and pattern recognition},
  pages={9527--9537},
  year={2023}
}

@inproceedings{huang2024alignsam,
  title={Alignsam: Aligning segment anything model to open context via reinforcement learning},
  author={Huang, Duojun and Xiong, Xinyu and Ma, Jie and Li, Jichang and Jie, Zequn and Ma, Lin and Li, Guanbin},
  booktitle={Proceedings of the IEEE/CVF conference on computer vision and pattern recognition},
  pages={3205--3215},
  year={2024}
}

@inproceedings{zhao2024lake,
  title={Lake-red: Camouflaged images generation by latent background knowledge retrieval-augmented diffusion},
  author={Zhao, Pancheng and Xu, Peng and Qin, Pengda and Fan, Deng-Ping and Zhang, Zhicheng and Jia, Guoli and Zhou, Bowen and Yang, Jufeng},
  booktitle={Proceedings of the IEEE/CVF Conference on Computer Vision and Pattern Recognition},
  pages={4092--4101},
  year={2024}
}

@inproceedings{wang2022ease,
  title={Ease: Robust facial expression recognition via emotion ambiguity-sensitive cooperative networks},
  author={Wang, Lijuan and Jia, Guoli and Jiang, Ning and Wu, Haiying and Yang, Jufeng},
  booktitle={Proceedings of the 30th ACM international conference on multimedia},
  pages={218--227},
  year={2022}
}

@inproceedings{yang2024event,
  title={Event-based Few-shot Fine-grained Human Action Recognition},
  author={Yang, Zonglin and Yang, Yan and Shi, Yuheng and Yang, Hao and Zhang, Ruikun and Liu, Liu and Wu, Xinxiao and Pan, Liyuan},
  booktitle={2024 IEEE/RSJ International Conference on Intelligent Robots and Systems (IROS)},
  pages={519--526},
  year={2024},
  organization={IEEE}
}

@article{pan2020high,
  title={High frame rate video reconstruction based on an event camera},
  author={Pan, Liyuan and Hartley, Richard and Scheerlinck, Cedric and Liu, Miaomiao and Yu, Xin and Dai, Yuchao},
  journal={IEEE Transactions on Pattern Analysis and Machine Intelligence},
  volume={44},
  number={5},
  pages={2519--2533},
  year={2020},
  publisher={IEEE}
}

@InProceedings{Yang_2025_CVPR,
    author    = {Yang, Yan and Pan, Liyuan and Li, Dongxu and Liu, Liu},
    title     = {EZSR: Event-based Zero-Shot Recognition},
    booktitle = {Proceedings of the IEEE/CVF Conference on Computer Vision and Pattern Recognition (CVPR)},
    month     = {June},
    year      = {2025},
    pages     = {4628-4638}
}

@inproceedings{pan2020single,
  title={Single image optical flow estimation with an event camera},
  author={Pan, Liyuan and Liu, Miaomiao and Hartley, Richard},
  booktitle={2020 IEEE/CVF Conference on Computer Vision and Pattern Recognition (CVPR)},
  pages={1669--1678},
  year={2020},
  organization={IEEE}
}

@InProceedings{Gallego_2018_CVPR,
author = {Gallego, Guillermo and Rebecq, Henri and Scaramuzza, Davide},
title = {A Unifying Contrast Maximization Framework for Event Cameras, With Applications to Motion, Depth, and Optical Flow Estimation},
booktitle = {Proceedings of the IEEE Conference on Computer Vision and Pattern Recognition (CVPR)},
month = {June},
year = {2018}
}

@inproceedings{hu2024robust,
  title={Robust facial reactions generation: An emotion-aware framework with modality compensation},
  author={Hu, Guanyu and Wei, Jie and Song, Siyang and Kollias, Dimitrios and Yang, Xinyu and Sun, Zhonglin and Kaloidas, Odysseus},
  booktitle={2024 IEEE International Joint Conference on Biometrics (IJCB)},
  pages={1--10},
  year={2024},
  organization={IEEE}
}

@inproceedings{hazarika2018icon,
  title={Icon: Interactive conversational memory network for multimodal emotion detection},
  author={Hazarika, Devamanyu and Poria, Soujanya and Mihalcea, Rada and Cambria, Erik and Zimmermann, Roger},
  booktitle={Proceedings of the 2018 conference on empirical methods in natural language processing},
  pages={2594--2604},
  year={2018}
}

@article{poria2019emotion,
  title={Emotion recognition in conversation: Research challenges, datasets, and recent advances},
  author={Poria, Soujanya and Majumder, Navonil and Mihalcea, Rada and Hovy, Eduard},
  journal={IEEE access},
  volume={7},
  pages={100943--100953},
  year={2019},
  publisher={IEEE}
}

@inproceedings{wei2023multi,
  title={Multi-scale receptive field graph model for emotion recognition in conversations},
  author={Wei, Jie and Hu, Guanyu and Tuan, Luu Anh and Yang, Xinyu and Zhu, Wenjing},
  booktitle={ICASSP 2023-2023 IEEE International Conference on Acoustics, Speech and Signal Processing (ICASSP)},
  pages={1--5},
  year={2023},
  organization={IEEE}
}

@inproceedings{zheng2023facial,
  title={A facial expression-aware multimodal multi-task learning framework for emotion recognition in multi-party conversations},
  author={Zheng, Wenjie and Yu, Jianfei and Xia, Rui and Wang, Shijin},
  booktitle={Proceedings of the 61st Annual Meeting of the Association for Computational Linguistics (Volume 1: Long Papers)},
  pages={15445--15459},
  year={2023}
}

@inproceedings{hu2022mm,
  title={MM-DFN: Multimodal dynamic fusion network for emotion recognition in conversations},
  author={Hu, Dou and Hou, Xiaolong and Wei, Lingwei and Jiang, Lianxin and Mo, Yang},
  booktitle={ICASSP 2022-2022 IEEE International Conference on Acoustics, Speech and Signal Processing (ICASSP)},
  pages={7037--7041},
  year={2022},
  organization={IEEE}
}

@article{ekman1999basic,
  title={Basic emotions},
  author={Ekman, Paul and Dalgleish, Tim and Power, M},
  journal={San Francisco, USA},
  volume={1},
  year={1999}
}

@inproceedings{mittal2020m3er,
  title={M3er: Multiplicative multimodal emotion recognition using facial, textual, and speech cues},
  author={Mittal, Trisha and Bhattacharya, Uttaran and Chandra, Rohan and Bera, Aniket and Manocha, Dinesh},
  booktitle={Proceedings of the AAAI conference on artificial intelligence},
  volume={34},
  number={02},
  pages={1359--1367},
  year={2020}
}

@inproceedings{sikka2013multiple,
  title={Multiple kernel learning for emotion recognition in the wild},
  author={Sikka, Karan and Dykstra, Karmen and Sathyanarayana, Suchitra and Littlewort, Gwen and Bartlett, Marian},
  booktitle={Proceedings of the 15th ACM on International conference on multimodal interaction},
  pages={517--524},
  year={2013}
}

@article{schlosberg1954three,
  title={Three dimensions of emotion.},
  author={Schlosberg, Harold},
  journal={Psychological review},
  volume={61},
  number={2},
  pages={81},
  year={1954},
  publisher={American Psychological Association}
}

@inproceedings{jiang2014predicting,
  title={Predicting emotions in user-generated videos},
  author={Jiang, Yu-Gang and Xu, Baohan and Xue, Xiangyang},
  booktitle={Proceedings of the AAAI conference on artificial intelligence},
  volume={28},
  number={1},
  year={2014}
}

@article{xu2016heterogeneous,
  title={Heterogeneous knowledge transfer in video emotion recognition, attribution and summarization},
  author={Xu, Baohan and Fu, Yanwei and Jiang, Yu-Gang and Li, Boyang and Sigal, Leonid},
  journal={IEEE Transactions on Affective Computing},
  volume={9},
  number={2},
  pages={255--270},
  year={2016},
  publisher={IEEE}
}

@inproceedings{hu2021v2e,
  title={v2e: From video frames to realistic DVS events},
  author={Hu, Yuhuang and Liu, Shih-Chii and Delbruck, Tobi},
  booktitle={Proceedings of the IEEE/CVF conference on computer vision and pattern recognition},
  pages={1312--1321},
  year={2021}
}

@article{shi2026qwen3,
  title={Qwen3-ASR Technical Report},
  author={Shi, Xian and Wang, Xiong and Guo, Zhifang and Wang, Yongqi and Zhang, Pei and Zhang, Xinyu and Guo, Zishan and Hao, Hongkun and Xi, Yu and Yang, Baosong and others},
  journal={arXiv preprint arXiv:2601.21337},
  year={2026}
}

@article{yu2024yolo,
  title={Yolo-facev2: A scale and occlusion aware face detector},
  author={Yu, Ziping and Huang, Hongbo and Chen, Weijun and Su, Yongxin and Liu, Yahui and Wang, Xiuying},
  journal={Pattern Recognition},
  volume={155},
  pages={110714},
  year={2024},
  publisher={Elsevier}
}

@article{kartynnik2019real,
  title={Real-time facial surface geometry from monocular video on mobile GPUs},
  author={Kartynnik, Yury and Ablavatski, Artsiom and Grishchenko, Ivan and Grundmann, Matthias},
  journal={arXiv preprint arXiv:1907.06724},
  year={2019}
}

@inproceedings{baltrusaitis2018openface,
  title={OpenFace 2.0: Facial behavior analysis toolkit},
  author={Baltru{\v{s}}aitis, Tadas and Zadeh, Amir and Lim, Yao Chong and Morency, Louis-Philippe},
  booktitle={2018 13th IEEE international conference on automatic face \& gesture recognition (FG 2018)},
  pages={59--66},
  year={2018},
  organization={IEEE}
}

@inproceedings{hou2025masked,
  title={Masked Image Pretraining on Language Assisted Representation},
  author={Hou, Zejiang and Kung, Sun-Yuan},
  booktitle={ICASSP 2025-2025 IEEE International Conference on Acoustics, Speech and Signal Processing (ICASSP)},
  pages={1--5},
  year={2025},
  organization={IEEE}
}

@inproceedings{zhang2024mart,
  title={Mart: Masked affective representation learning via masked temporal distribution distillation},
  author={Zhang, Zhicheng and Zhao, Pancheng and Park, Eunil and Yang, Jufeng},
  booktitle={Proceedings of the IEEE/CVF conference on computer vision and pattern recognition},
  pages={12830--12840},
  year={2024}
}

@inproceedings{huang2023mgmae,
  title={Mgmae: Motion guided masking for video masked autoencoding},
  author={Huang, Bingkun and Zhao, Zhiyu and Zhang, Guozhen and Qiao, Yu and Wang, Limin},
  booktitle={Proceedings of the IEEE/CVF International Conference on Computer Vision},
  pages={13493--13504},
  year={2023}
}

@inproceedings{chen2021empirical,
  title={An empirical study of training self-supervised vision transformers},
  author={Chen, Xinlei and Xie, Saining and He, Kaiming},
  booktitle={Proceedings of the IEEE/CVF international conference on computer vision},
  pages={9640--9649},
  year={2021}
}

@article{liu2019roberta,
  title={Roberta: A robustly optimized bert pretraining approach},
  author={Liu, Yinhan and Ott, Myle and Goyal, Naman and Du, Jingfei and Joshi, Mandar and Chen, Danqi and Levy, Omer and Lewis, Mike and Zettlemoyer, Luke and Stoyanov, Veselin},
  journal={arXiv preprint arXiv:1907.11692},
  year={2019}
}

@article{hsu2021hubert,
  title={Hubert: Self-supervised speech representation learning by masked prediction of hidden units},
  author={Hsu, Wei-Ning and Bolte, Benjamin and Tsai, Yao-Hung Hubert and Lakhotia, Kushal and Salakhutdinov, Ruslan and Mohamed, Abdelrahman},
  journal={IEEE/ACM transactions on audio, speech, and language processing},
  volume={29},
  pages={3451--3460},
  year={2021},
  publisher={IEEE}
}

@inproceedings{Zhao2020AnEV,
  title={An End-to-End Visual-Audio Attention Network for Emotion Recognition in User-Generated Videos},
  author={Sicheng Zhao and Yunsheng Ma and Yang Gu and Jufeng Yang and Tengfei Xing and Pengfei Xu and Runbo Hu and Hua Chai and Kurt Keutzer},
  booktitle={AAAI},
  year={2020}
}

@INPROCEEDINGS{9102808,
  author={Qiu, Haonan and He, Liang and Wang, Feng},
  booktitle={2020 IEEE International Conference on Multimedia and Expo (ICME)}, 
  title={Dual Focus Attention Network For Video Emotion Recognition}, 
  year={2020},
  volume={},
  number={},
  pages={1-6},
  doi={10.1109/ICME46284.2020.9102808}}

@inproceedings{you2025lvface,
  title={LVFace: Progressive cluster optimization for large vision models in face recognition},
  author={You, Jinghan and Li, Shanglin and Sun, Yuanrui and Wei, Jiangchuan and Guo, Mingyu and Feng, Chao and Ran, Jiao},
  booktitle={Proceedings of the IEEE/CVF International Conference on Computer Vision},
  pages={11840--11849},
  year={2025}
}
\end{document}